\documentclass[nofootinbib,superscriptaddress,a4paper,twocolumn]{revtex4-1}

\usepackage{array,mathtools,amssymb,booktabs}
\newcolumntype{C}{>{$}c<{$}}
\AtBeginDocument{
\heavyrulewidth=.08em
\lightrulewidth=.05em
\cmidrulewidth=.03em
\belowrulesep=.65ex
\belowbottomsep=0pt
\aboverulesep=.4ex
\abovetopsep=0pt
\cmidrulesep=\doublerulesep
\cmidrulekern=.5em
\defaultaddspace=.5em
}
\usepackage{hyperref}

\usepackage{geometry}
\usepackage[english]{babel}
\usepackage[latin1]{inputenc}
\usepackage{hyperref}
\usepackage{amsfonts}
\usepackage{amsmath, amssymb, amsfonts, mathrsfs}
\usepackage{graphicx}
\usepackage{xspace}
\usepackage{bbm}
\usepackage{natbib}
\usepackage{multirow}
\usepackage[bottom]{footmisc}
\usepackage{array}
\usepackage{tabulary}
\usepackage{booktabs}
\usepackage{enumerate}
\newcommand\mydots{\hbox to 1em{.\hss.\hss.}}

\newcommand\blfootnote[1]{%
  \begingroup
  \renewcommand\thefootnote{}\footnote{#1}%
  \addtocounter{footnote}{-1}%
  \endgroup
}

\newcommand*{\selfies}{{S}{\footnotesize ELFIES}\xspace}

\begin{document}

\title{Self-Referencing Embedded Strings (SELFIES):\\A 100\% robust molecular string representation}
\author{Mario Krenn}
\email{mario.krenn@utoronto.ca}
\affiliation{Department of Chemistry, University of Toronto, Canada.}
\affiliation{Department of Computer Science, University of Toronto, Canada.}
\affiliation{Vector Institute for Artificial Intelligence, Toronto, Canada.}
\author{Florian H\"ase}
\affiliation{Department of Chemistry, University of Toronto, Canada.} 
\affiliation{Department of Computer Science, University of Toronto, Canada.}
\affiliation{Vector Institute for Artificial Intelligence, Toronto, Canada.}
\affiliation{Department of Chemistry and Chemical Biology, Harvard University, Cambridge, USA.}

\author{AkshatKumar Nigam}
\affiliation{Department of Computer Science, University of Toronto, Canada.} 

\author{Pascal Friederich}
\affiliation{Department of Computer Science, University of Toronto, Canada.}
\affiliation{Institute of Nanotechnology, Karlsruhe Institute of Technology, Germany.}

\author{Alan Aspuru-Guzik}
\email{alan@aspuru.com}
\affiliation{Department of Chemistry, University of Toronto, Canada.}
\affiliation{Department of Computer Science, University of Toronto, Canada.}
\affiliation{Vector Institute for Artificial Intelligence, Toronto, Canada.}
\affiliation{Canadian Institute for Advanced Research (CIFAR) Senior Fellow, Toronto, Canada.}

\begin{abstract}
The discovery of novel materials and functional molecules can help to solve some of society's most urgent challenges, ranging from efficient energy harvesting and storage to uncovering novel pharmaceutical drug candidates. Traditionally matter engineering -- generally denoted as inverse design -- was based massively on human intuition and high-throughput virtual screening. The last few years have seen the emergence of significant interest in computer-inspired designs based on evolutionary or deep learning methods. The major challenge here is that the standard strings molecular representation SMILES shows substantial weaknesses in that task because large fractions of strings do not correspond to valid molecules. Here, we solve this problem at a fundamental level and introduce \selfies (SELF-referencIng Embedded Strings), a string-based representation of molecules which is 100\% robust. Every \selfies string corresponds to a valid molecule, and \selfies can represent every molecule. \selfies can be directly applied in arbitrary machine learning models without the adaptation of the models; each of the generated molecule candidates is valid. In our experiments, the model's internal memory stores two orders of magnitude more diverse molecules than a similar test with SMILES. Furthermore, as all molecules are valid, it allows for explanation and interpretation of the internal working of the generative models.
\end{abstract}
\date{\today}
\maketitle

\textbf{Introduction}\blfootnote{The full code is available at \texttt{GitHub}: \url{https://github.com/aspuru-guzik-group/selfies}} -- The rise of computers enabled the creation of the field of computational chemistry and cheminformatics which deals with the development and application of methods to calculate, process, store and search molecular information on computing systems. Arising challenges of molecular representation and identification were addressed by SMILES (Simplified Molecular Input Line Entry System), which was invented by David Weiniger in 1988 \cite{weininger1988smiles}. SMILES is a simple string-based representation which is based on principles of molecular graph theory and allows molecular structure specification with straightforward rules. SMILES has since become a standard tool in computational chemistry and is still a de-facto standard for string-based representing molecular information in-silico.

\begin{figure}[!b]
\centering
\includegraphics[width=0.4\textwidth]{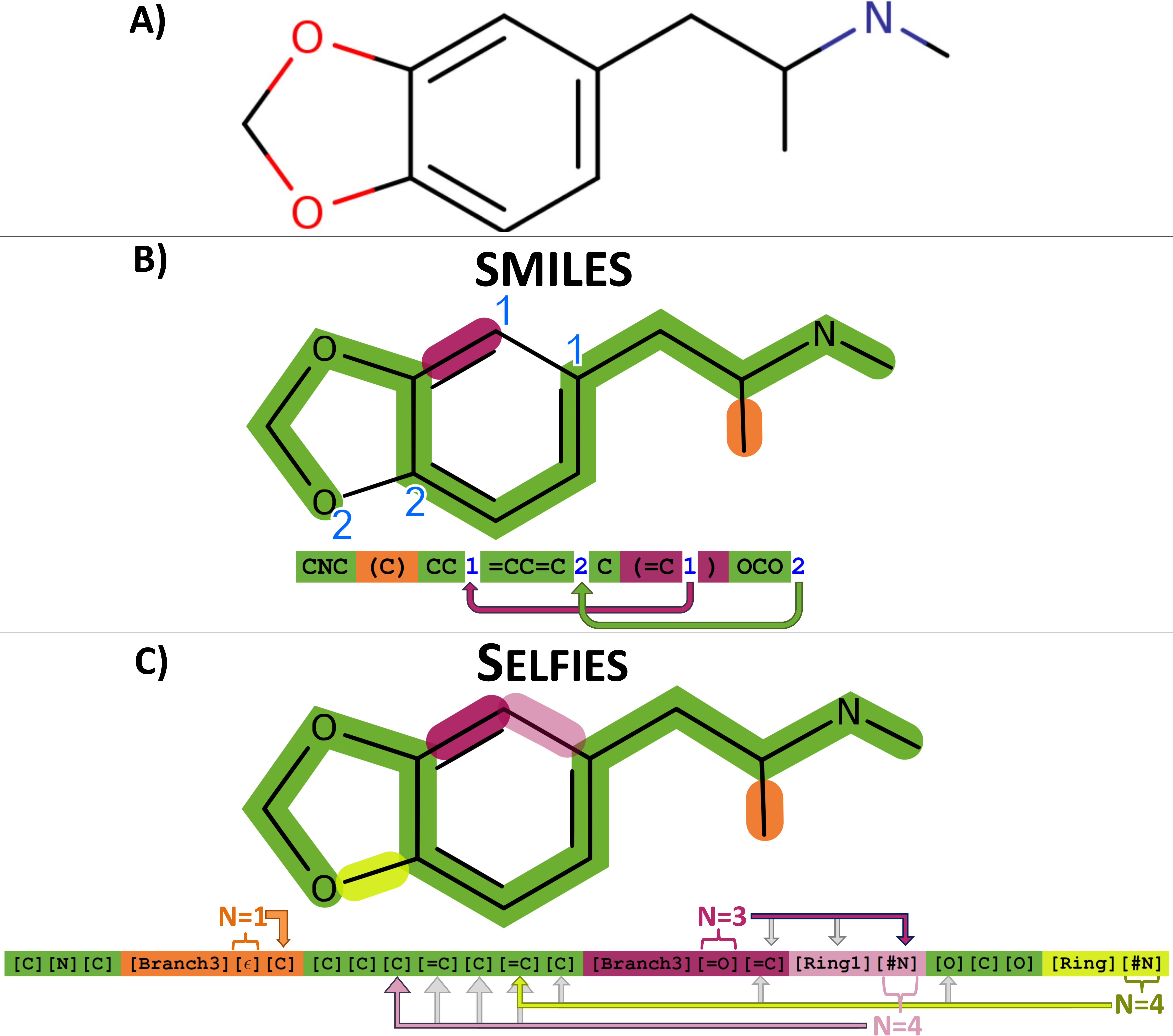}
\caption{Description of a molecular graph with two computer-friendly, string-based methods. \textbf{A)} The molecular graph of a small organic molecule, 3,4-Methylenedioxymethamphetamine.  \textbf{B)} Derivation of the molecular graph using SMILES. The main string (green) is augmented with branches (defined by an opening and a closing bracket) and rings (defined by unique numbers after the atoms that are connected). Note that both branches and rings are non-local operations. \textbf{C)} Derivation of the molecular graph using \selfies. The main string is derived using a rule set such that the number of valence bonds per atom does not exceed physical limits. The symbol after a \texttt{Branch} is interpreted as the number of \selfies symbols derived inside the branch. The symbol after \texttt{Ring} interpreted as a number too, indicating that the current atom is connected to the ($N+1$)st previous atom. Thereby every information in the string (except the ring closure) is local and allows for efficient derivation rules.}
\label{fig:mdma}
\end{figure}

\begin{figure*}[!t]
\centering
\includegraphics[width=0.95\textwidth]{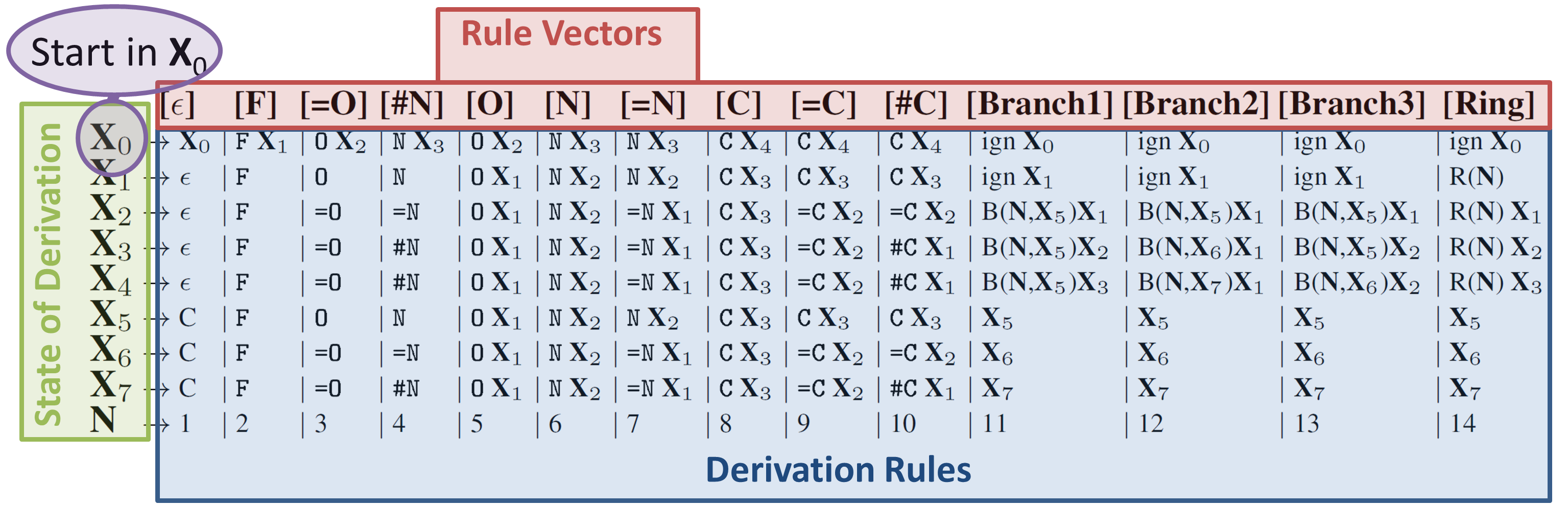}
\caption{Derivation rules of \selfies for small organic molecules. Every symbol of \selfies is interpreted as a rule vector (top red line). A \selfies symbol will be replaced by the string at the intersection of the rule vector and derivation state of the derivation (left, green). The string can contain an atom or another state of derivation. The derivation starts in the state $\mathbf{X}_0$ (violet), and continues in the state previously derived. The state of derivation takes care of syntactical and chemical constraints, such as the maximal number of valence bonds. The rules in state $\mathbf{X}_n$ for $n=1$-$n=4$ are designed such the next atom can use up to $n$ valence bonds. $B(N,\mathbf{X_n})$ stands for function, creating a branch in the graph using the next $N$ symbols and starting in state $\mathbf{X_n}$. $R(N)$ stands for a function that creates rings, from the current atom to the ($N+1$)-st previously derived atom. In both cases, the letter subsequent to $R$ or $B$ is interpreted as a number $N$, which is defined in the last line of the table. This table covers all non-ionic molecules in the database QM9 \cite{ramakrishnan2014quantum, ruddigkeit2012enumeration}. Ions, stereochemistry and larger molecules can also be represented by simply extending this table, as we show in the SI. } 
\label{fig:table}
\end{figure*}

Apart from predicting molecular properties with high accuracy, one of the main goals in computational chemistry is the design of novel, functional molecules. Exploring the entire chemical space -- even for relatively small molecules -- is intractable due to the combinatorial explosion of possible and stable chemical structures \cite{oprea2001chemography,virshup2013stochastic,qian2015exploring}. Substantial recent advances in artificial intelligence and machine learning (ML), in particular, the development and control of generative models, have found their way into chemical research. There, scientists are currently adapting those novel methods for efficiently proposing new molecules with superior properties \cite{raccuglia2016machine,sanchez2018inverse,jorgensen2018deep, elton2019deep, gromski2019explore, jensen2019graph}. For identifying new molecules, input and output representations are in many cases SMILES strings. This, however, introduces a substantial problem: A significant fraction of the resulting SMILES strings do not correspond to valid molecules. They are either syntactically invalid, i.e. do not even correspond to a molecular graph, or they violate basic chemical rules, such as the maximum number of valence bonds between atoms. Researchers have proposed many special-case solutions for overcoming these problems, (such as adapting specific machine learning models \cite{ma2018constrained, liu2018constrained} or changing some definitions of SMILES \cite{dalke2018deepsmiles}), however, a universal solution is lacking. Thus, more than 30 years after Weininger's invention of SMILES, the applications of generative models for the de-novo design of molecules requires a new way to describe molecules on the computer.

Here, we present \selfies (SELF-referencIng Embedded Strings), a string-based representation of molecular graphs that is 100\% robust. By that, we mean that each \selfies corresponds to a valid molecule, even entirely random strings. Furthermore, every molecule can be described as a \selfies. \selfies are independent of the machine learning model and can be used as a direct input without any adaptations of the models.

We compare \selfies with SMILES ML-based generative models such as in Variational Autoencoders (VAE) \cite{kingma2013auto} and Generative Adversarial Networks (GANs) \cite{goodfellow2014generative}. We find that the output is entirely valid and the models encode orders of magnitude more diverse molecules with \selfies than with SMILES. Those results are not only significant for inverse-design of molecules, but also interpretability of the inner workings of neural networks in the chemical domain.

\textbf{String-based representations of Molecules --}      
We are describing the string-based representations of SMILES and \selfies using the small biomolecule 3,4-Methylenedioxymethamphetamine (MDMA) in Fig. \ref{fig:mdma}A. The SMILES string in Fig. \ref{fig:mdma}B describes a sequence of connected atoms (green). Brackets identify branches and, and numbers identify ring-closures at the atoms that are connected. In \selfies, Fig. \ref{fig:mdma}C, the information of branch length as well as ring size is stored together with the corresponding identifiers \texttt{Branch} and \texttt{Ring}. For that, the symbol after the \texttt{Branch} and \texttt{Ring} stands for a number that is interpreted as lengths. Thereby, the possibility of invalid syntactical string (such as a string with more opening than closing brackets), is prevented. Furthermore, each \selfies symbols is generated using derivation rules, see Table \ref{fig:table}. Formally, the table corresponds to a formal grammar from theoretical computer science \cite{Hopcroft}. The derivation of a single symbol depends on the state of the derivation $\mathbf{X_n}$. The purpose of these rules is to enforce the validity of the chemical valence-bonds.

As a simple example, the string \texttt{[F][=C][=C][\#N]} is derived in the following way. Starting in the state $\mathbf{X_0}$, the first symbol (rule vector) \texttt{[F]} leads to \texttt{F} $\mathbf{X_1}$. The derivation of the second symbol subsequently continues in the state $\mathbf{X_1}$. The total derivation is given by
\begin{align}
\mathbf{X_0}&\xmapsto{[F]}\texttt{F} \mathbf{X_1} \xmapsto{[=C]} \texttt{FC} \mathbf{X_3} \nonumber\\
&\xmapsto{[=C]} \texttt{FC=C} \mathbf{X_2} \xmapsto{[\#N]} \texttt{FC=C=N} \nonumber
\end{align}

The final molecule \texttt{FC=C=N}, which satisfies all valence-bond rules, is 2-Fluoroethenimine. At this point, valence-bond constraints are satisfied for subsequent atoms and branches. The only remaining potential sources of violation of these constraints are the destination of rings. Therefore, we insert rings only if the number of valence-bond at the target has not yet reached the maximum. Thereby, using the rules in Table \ref{fig:table}, 100\% validity can be guaranteed for small biomolecules. It is straight forward to extend the coverage for broader classes of molecules, as we describe below.

The derivation rules in Table \ref{fig:table} are generated systematically and could be constructed fully automatically just from data, as we show in the SI. Furthermore, \selfies are not restricted to molecular graphs but could be applied to other graph data types in the natural sciences that have additional domain-dependent constraints. We give an example, quantum optical experiments in physics with component dependent connectivity \cite{krenn2016automated}, in the SI.

Informal conversations with several researchers lead to the argument that SMILES are "readable". Readability is in the eye of the beholder, but needless to say, \selfies are as readable as figure \ref{fig:mdma}C) attests to. After a little familiarity, functional groups and connectivity can be inferred by human interpretation for small molecular fragments.
\begin{figure}[!t]
\centering
\includegraphics[width=0.5\textwidth]{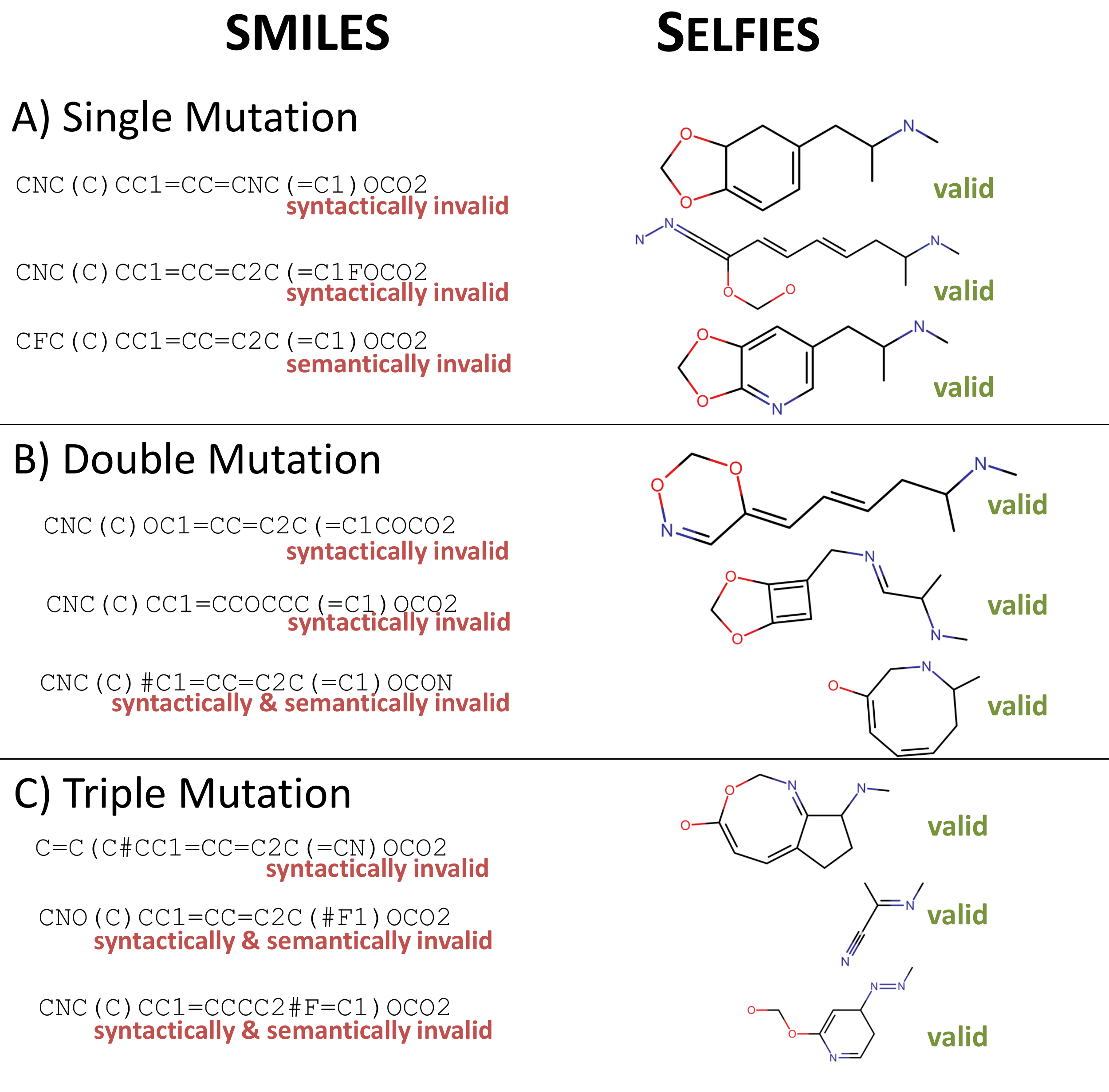}
\caption{Random Mutations of SMILES and \selfies of the molecule in Fig. \ref{fig:mdma}A. A) Single mutations have led to three invalid SMILES strings, while all \selfies produce valid molecules. In B) and C) the initial molecule is two and three times mutated, respectively. In all cases, SMILES strings are invalid, while \selfies produce valid molecules that deviate more and more from the initial molecule.}
\label{fig:mutations}
\end{figure}

\textbf{Effects of random Mutations --} The simplest way to compare robustness between SMILES and \selfies is by starting from a valid string, such as MDMA in Fig. \ref{fig:mdma}, and introduce random mutations of the symbols of the string. In Fig. \ref{fig:mutations}A, we show three examples of one randomly introduced string mutation. We evaluate the resulting validity using RDKit \cite{landrum2006rdkit}. All three SMILES strings are invalid. The first one is missing a second ring-identifier for \texttt{2}, the second one is missing a closing bracket for a branch, and the last one violates valence-bond numbers of Fluorine. In contrast to that, all mutated \selfies correspond to valid molecules. In Fig.\ref{fig:mutations}B and C, we introduce two and three mutations, respectively. Again, all SMILES are invalid, and all \selfies are valid molecules. In general, the validity probability for SMILES with one mutation starting from MDMA is 9.9\%, 3.0\% and 1.1\% for one, two and three mutations, respectively. \selfies, on the other hand, are valid in 100\% of the cases. Three examples for each case can be seen on the right panel of Fig. \ref{fig:mutations}. \footnote{We also investigate the validity rates of a recent adaption of SMILES denoted DeepSMILES \cite{dalke2018deepsmiles}. DeepSMILES could also be used as a direct input for arbitrary machine learning models and follows, therefore, a similar objective as \selfies. We find that single, double and triple mutations lead to 35.1\%, 18.4\% and 9.8\% validity.}

\begin{figure}[!t]
\centering
\includegraphics[width=0.45\textwidth]{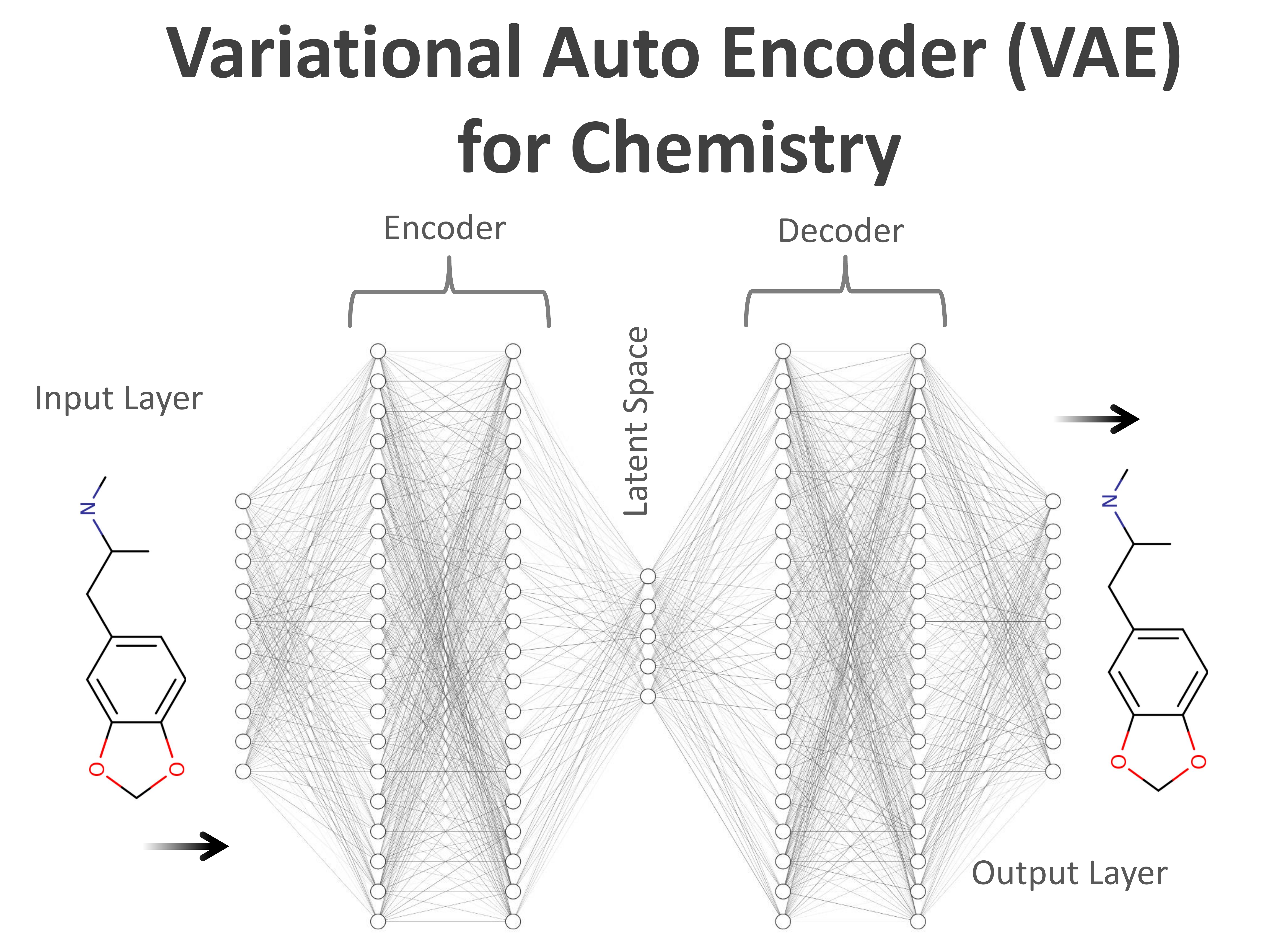}
\caption{Variational Autoencoder (VAE) for Chemistry. The VAE is a deep neural network that takes a molecule as an input, encodes it to continuous latent space, and reconstructs it from there with a decoder. The latent space is a high-dimensional space where each point can be decoded into a discrete sequence. We represent the molecular graphs using one-hot encodings of SMILES and \selfies.}
\label{fig:VAE}
\end{figure}

\textbf{Results for deep generative Models --} Generative models are an ideal application of a 100\% robust representation of molecules. One prominent example is a variational autoencoder (VAE) \cite{kingma2013auto}, which has recently been employed for the design of novel molecules \cite{gomez2018automatic}. In the domain of chemistry, the VAE is used to transform a discrete molecular graph into a continuous representation which can be optimized using gradient-based or Bayesian methods. As shown in Fig. \ref{fig:VAE}, it consists of two neural networks, the encoder and decoder. The encoder takes a string representation of the molecule (for instance, using one-hot encoding) and encodes it into a continuous, internal representation. There, every molecule corresponds to a location in a high-dimensional space. The number of neurons defines the dimension in the latent space. The decoder takes a position in the latent space and transforms it into a discrete molecule (for instance again, a one-hot encoding of SMILES or \selfies).

\begin{figure}[!b]
\centering
\includegraphics[width=0.48\textwidth]{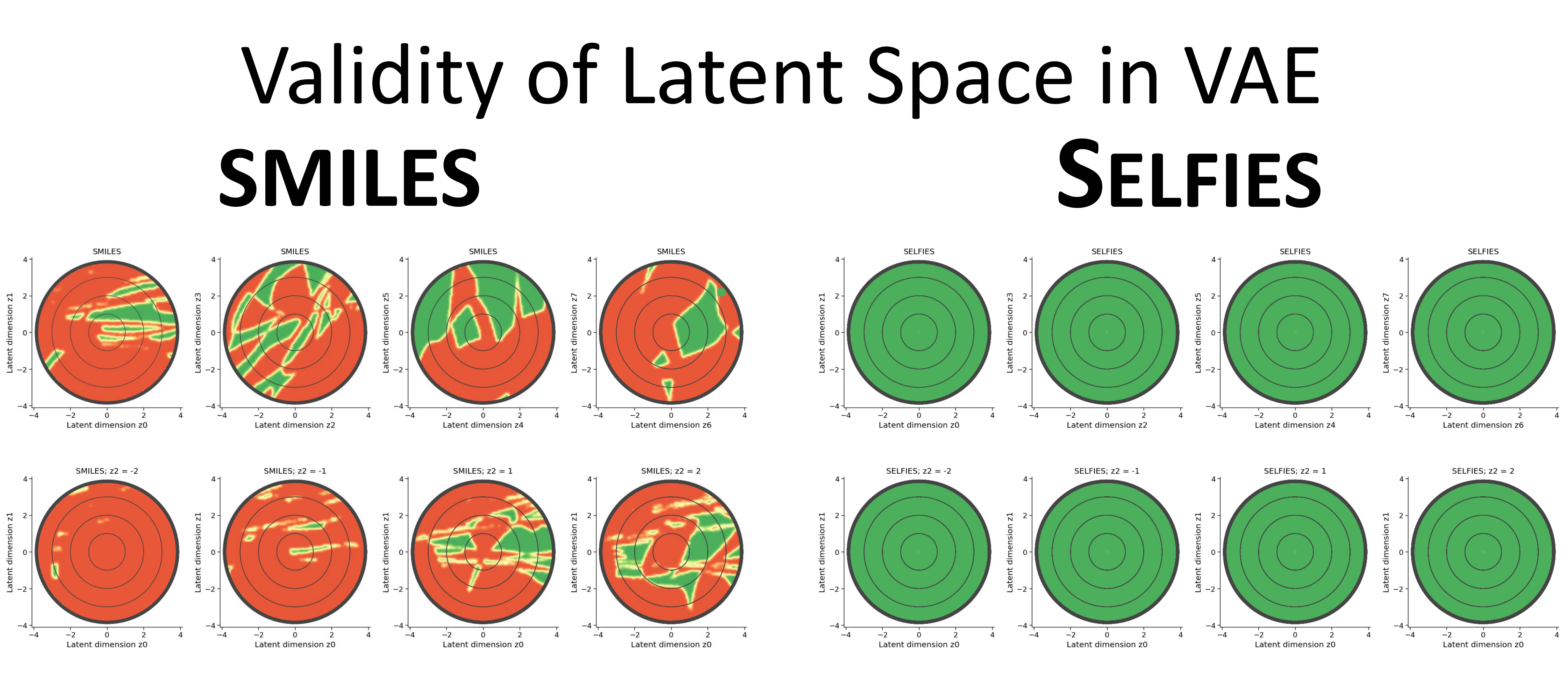}
\caption{Validity of latent space. We analyze the latent space of a VAE, which was trained to reproduce small organic molecules from the QM9 database. The latent space has 241 dimensions (LD stands for latent dimension). Upper row: We chose four randomly oriented planes in the high-dimensional space that go through the origin. Along the plane, we decode latent space points and calculate whether they correspond to valid or invalid molecules. The color code stands for the proportion of valid molecules (red=0\%, green=100\% valid). Lower row: We chose a random orientation of the plane, and displace it by a third random orientation by (-2,-1,+1,+2) standard deviations from the origin. In all experiments, we find that only a small fraction of the latent space for SMILES are valid, while for \selfies the entire latent space is valid. This is not only important for generative tasks but is crucial for interpreting internals representations of the neural networks.}
\label{fig:validity}
\end{figure}

\begin{figure*}
\centering
\includegraphics[width=0.7\textwidth]{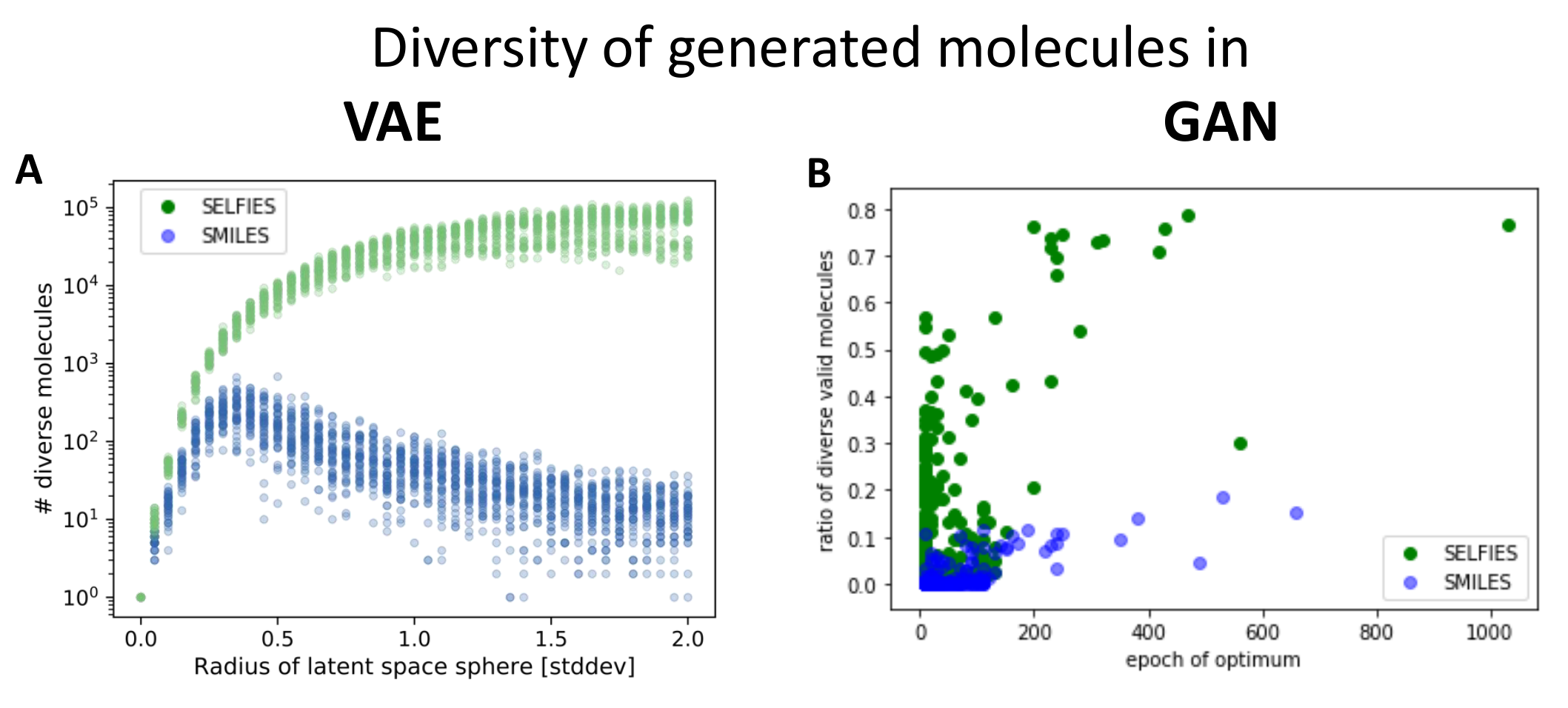}
\caption{Diversity of generative models trained with SMILES and \selfies, with the example of VAE and GAN. Beside robustness, diversity is one of the main objectives for generative models. A) We investigate the density of valid diverse molecules by sampling the latent space of a VAE. We chose points with a distance of $\sigma$ around the centre, stopping after 20 samples didn't produce new instances. We find that the VAE trained with \selfies contains 100 times more valid diverse molecules than if it is trained with SMILES. B) We train a GAN with 200 different hyperparameters to produce de-novo molecules for \selfies and SMILES. Sampling 10.000 times, \selfies produced 7889 different valid molecules, while for SMILES the most diverse valid number of molecules where 1855). Both cases show that \selfies leads to significantly larger densities of diverse molecules compared to SMILES.}
\label{fig:diversity}
\end{figure*}

The goal of a VAE is learning to reconstruct molecules. After the training, one can scan through the latent space for optimizing chemical properties. Once an optimal point is identified, the decoder can map it to a molecular string. For any application of VAEs in chemistry, it is desirable that all points in the latent space correspond to valid molecules.

We experiment with a standard VAE, which we train to reconstruct molecules from the benchmark dataset QM9 \cite{ramakrishnan2014quantum, ruddigkeit2012enumeration}. We employ both SMILES and \selfies for that task. After the training, we analyze the validity of the latent space. We do this by sampling latent space points from randomly oriented planes in the high-dimensional space. Using SMILES, we find in Fig. \ref{fig:validity}A that only a small fraction of the space corresponds to valid molecules. A large fraction decodes to syntactically or semantically invalid strings that do not stand for molecules. In contrast to that, using \selfies, we can see in Fig.\ref{fig:validity}B that the entire space corresponds to valid molecules. We want to stress that a 100\% valid latent space is not only significant for inverse-design techniques in chemistry, but is essential for model interpretation \cite{schutt2017quantum, preuer2019interpretable, hase2019machine}, in particular for interpreting the internal representations \cite{higgins2017beta, chen2018isolating} in a scientific context \cite{iten2020discovering}.

Besides 100\% validity, the molecule density in the latent space is of crucial importance too. The more valid, diverse molecules are encoded inside the latent space, the richer the chemical space that can be explored during optimization procedures. In Fig. \ref{fig:diversity}A, we compare the richness of the encoded molecules when a VAE is trained with SMILES and with \selfies. For that, we sample random points in the latent space and stop after 20 samples didn't produce any new molecule. We find that the latent space of the \selfies VAE is more than two orders of magnitude denser than the one of SMILES.

Other prominent deep generative models are Generative Adversarial Networks (GANs) \cite{goodfellow2014generative}, which have been introduced in the design of molecules \cite{BenORGAN}. There, two networks -- called generator and discriminator -- are trained in tandem. The setting is such that discriminator receives either molecule from a dataset or outputs of the generator. The goal of the discriminator is to correctly identify the artificially generated structures, while the goal of the generator is to fool the discriminator. After the training, the generator has learned to reproduce the distribution of the dataset. We train the GAN, using 200 different hyperparameter settings both for SMILES and \selfies. After the training, we sample each of the models 10.000 times and calculate the number of unique, valid molecules. For the best set of hyperparameters, we find that a GAN trained with \selfies produces 78.9\% diverse molecules while a GAN that produces SMILES strings only results in 18.6\% diverse molecules, see Fig. \ref{fig:diversity}B.

\textbf{Covering the chemical universe --} 
In this manuscript, we demonstrate and apply \selfies for small biomolecules. However, the language can be extended to cover much richer classes of molecules. In the corresponding \texttt{GitHub} repository, we extend the language to allow for molecules with up to 8000 atoms per ring and branch, we add stereochemistry information, ions as well as unconstrained unspecified symbols.  Thereby, we encoded and decoded all 72 million molecules from PubChem (the most complete collection of synthesized molecules) with less than 500 SMILES chars, demonstrating coverage of the space of chemical interest.

\textbf{Conclusion -- }
We presented \selfies, a human-readable and 100\% robust method to describe molecular graphs in a computer. These properties lead to superior behaviour in inverse design tasks for functional molecules, based on deep generative models or genetic algorithms. \selfies can be used as a direct input into current and even future generative models, without the requirement to adapt the model. In generative tasks, it leads to a significantly higher diversity of molecules, which is the main objective in inverse design. In addition to the results presented here, in separate work, we use Genetic Algorithms and find that without any hard-coded rules, \selfies outperform literature results in a commonly-used benchmark \cite{nigam2019augmenting}. Apart from superior behaviour in inverse design, a 100\% valid representation is also a sufficient condition for interpreting the internal structures of the machine learning models \cite{iten2020discovering}. While we have focused on an representation that is ideal for computers, attention should also be drawn to \selfies standardization to allow general readability \cite{o2018facto}, by exploiting the numerous remaining degrees of freedom of \selfies.

\textbf{Standarization outlook --}
The \selfies concept still requires work to become a standard. Upon publication of this article, the authors will call for a workshop to extend the format to the entire periodic table, allow for stereochemistry, polyvalency and other special cases so that all the features present in SMILES are available in selfies. Unicode will be employed to create readable symbols that exploit the flexibility of modern text systems without restricting oneself to ASCII characters.

\section*{Acknowledgements}
The authors thank Theophile Gaudin for useful discussions. A. A.-G. acknowledges generous support from the Canada 150 Research Chair Program, Tata Steel, Anders G. Froseth, and the Office of Naval Research. We acknowledge supercomputing support from SciNet. M.K. acknowledges support from the Austrian Science Fund (FWF) through the Erwin Schr\"odinger fellowship No. J4309. F.H. acknowledges support from the Herchel Smith Graduate Fellowship and the Jacques-Emile Dubois Student Dissertation Fellowship. P.F. has received funding from the European Union's Horizon 2020 research and innovation programme under the Marie Sklodowska-Curie grant agreement No 795206.
\bibliographystyle{unsrt}
\bibliography{refs}

\clearpage

\widetext
\begin{center}
\textbf{\Large Supplemental Materials}
\end{center}

\section{Formal definition of \selfies}
We take advantage of a \textbf{formal grammar} to derive \textit{words}, which will represent semantically valid graphs. A formal grammar is a tuple $G(V,\Sigma,R,S)$, where $v \in V$ are non-terminal symbols that are replaced using rules, $r \in R$, into non-terminal or terminal symbols $t \in \Sigma$. $S$ is a start symbol. When the resulting string only consists of terminal symbols, the derivation of a new word is completed \cite{Hopcroft}. The \selfies representation is a Chomsky type-2, context-free grammar with self-referencing functions for valid generation of branches in graphs. The rule system is shown in Table \ref{tab:GrammarAbstract}.

In \selfies, $V=\{\mathbf{X}_0, \dots, \mathbf{X}_r, \mathbf{N} \}$ are non-terminal symbols or states. The states $\mathbf{X}_i$ restrict the subsequent edge to a maximal multiplicity of $i$; the maximal edge multiplicity of the generated graphs is $r$. The symbol $\mathbf{N}$ represents a numerical value, which acts as argument for the two self-referencing functions. $\Sigma=\{t_{0,1}, \dots, t_{r,n}\}$ are terminal symbols. 
The derivation rule set $R$ has exactly $(n+m+p+1)\times(r+2)$ elements, corresponding to $n$ rules for vertex production, $m$ rules for producing branches, $p$ rules for rings and $r$ non-terminal symbols in $V$. The subscripts $h_{a,b}$, $i_{a,b}$, $j_{a,b}$ and $k_{a,b}$ have values from $1$ to $r$, and encode the actual domain-specific constraints. The semantic and syntactical constraints are encoded into the rule vectors, which guarantees strong robustness. There are $n+m+p+1$ rule vectors $\mathbf{A}_i$, each with a dimension $(r+2)$.  

\section{Self-referencing functions for syntactic validity}
In order to account for \textbf{syntactic validity} of the graph, we augment the context-free grammar with \textbf{branching functions} and \textbf{ring functions}. $B(\mathbf{N},X_i)$ is the branching function, that recursively starts another grammar derivation with subsequent $\mathbf{N}$ \selfies symbols in state $X_i$. After the full derivation of a new word (which is a graph), the branch function returns the graph, and connects it to the current vertex. The ring function $R(\mathbf{N})$ establishes edges between the current vertex and the $(\mathbf{N}+1)$-th last derived vertex. Both the branching and ring functions have access to the \selfies string and the derived string, thus are self-referencing. 

\section{Rule vectors for semantic validity}
To incorporate \textbf{semantic validity}, we denote $A_i$ as the \textit{i}-th vector of rules, with dimension $d_{A_i}=|V|=r+2$. The \textbf{conceptual idea} is to interpret a symbol of a \selfies string, $s_i \in \{0, \dots, n+m+p\}$ as an index of a rule vector, $A_{s_i}$. In the derivation of a symbol, the rule vector is defined by the symbol of the \selfies string (external state) while the specific rule is chosen by the non-terminal symbol (internal state). Thereby, we can encode semantic information into the rule vector $A_i$, which is \textit{memorized} by the internal state during derivation. 

\section{Algorithmic derivation of grammar from data, and validity guarantees}
Domain-specific grammars can be derived algorithmically directly from data, without any domain knowledge. Let $T$ be the set of different types of vertices (such as C, O, N, $\dots$ in chemistry). We use a dataset to get the types of vertices $T_i$, and their maximal degrees $D_i$ ($D_i=\mathrm{maxdeg}(T_i)$ -- in chemistry, the $D_{\mathrm{O}}=\mathrm{maxdeg}(\mathrm{O})=2$, $D_{\mathrm{C}}=\mathrm{maxdeg}(\mathrm{C})=4$). Let $M=\max_{i}\mathrm{maxdeg}(T_i)$ be the maximal degree of the dataset. Starting from Table \ref{tab:GrammarAbstract} (I) we identify the rule vectors $\mathbf{A}_i$, (II) define the non-terminal symbols $\mathbf{X}_j$, and (III) define the rules $R$. 
\begin{enumerate}[I]      
\item $\mathbf{A}_1 \dots \mathbf{A}_n$ (vertices rules) consist of $T_i$ with a potential multiedge connection $\gamma$ up to $D_i$ (in chemistry, $D_O$=2, thus we have two rule vectors for $O$, one with single edge $\gamma=1$, one with double connection $\gamma=2$). $\mathbf{A}_{n+1} \dots \mathbf{A}_{n+m}$ represent branch rules. A branch forms connections to two vertices, thus we have maximally ($M-1$) branch rules, (combinations of ($M-l,l$) represent the maximal connectivity to the two branches). $\mathbf{A}_{n+m+1} \dots \mathbf{A}_{n+m+p}$ denote ring rules, in a generic case $p=1$ is sufficient.
\item non-terminals $X_{1} \dots X_{r}$, with $r=M$, constrain the number of edges to connect two vertices.
\item Rule $r_{i,j}$ for $\mathbf{A}_i \in \{\mathbf{A}_1 \dots \mathbf{A}_n\}$ and $\mathbf{X}_j \in \{\mathbf{X}_{1} \dots \mathbf{X}_{r}\}$ can consist of a terminal and non-terminal symbol. The terminal consists of a $T_i$ (given by $\mathbf{A}_i$) and a edge-multiplicity $\mu=\min(j,\gamma)$. The corresponding nonterminal symbol is $\mathbf{X}_{M-\mu}$ (if $M-\mu=0$, no non-terminal will be added). Note that constraints are satisfied due to the $\min$ operation in $\mu$. Rules in state $\mathbf{X}_j$ for rings are $R(N)\mathbf{X}_{j-1}$, and for branches are $B(N,X_i)\mathbf{X}_{j-i}$. 
\end{enumerate}

\setlength{\tabcolsep}{1pt}
\begin{table*}[t]
    \small
    \centering
    \begin{tabular}{lllllllllllll}
        &&\multicolumn{4}{c}{ Vertices }&\multicolumn{3}{c}{ Branches }&\multicolumn{3}{c}{ Rings }\\         
        &&\multicolumn{4}{c}{$\overbrace{\hspace{40mm}}$}&\multicolumn{3}{c}{ $\overbrace{\hspace{50mm}}$}&\multicolumn{3}{c}{ $\overbrace{\hspace{33mm}}$}\\
        &&\multicolumn{1}{c}{\textbf{\large A}$_0$} & \multicolumn{1}{c}{\textbf{\large A}$_1$} && \multicolumn{1}{c}{\textbf{\large A}$_n$} & \multicolumn{1}{c}{\textbf{\large A}$_{n+1}$} && \multicolumn{1}{c}{\textbf{\large A}$_{n+m}$} & \multicolumn{1}{c}{\textbf{\large A}$_{n+m+1}$} && \multicolumn{1}{c}{\textbf{\large A}$_{n+m+p}$} \\               
        \textbf{\large X$_0$} & $\rightarrow$ & $\epsilon$ & $\vert$ $t_{0,1}$ \textbf{X$_{h_{0,1}}$}  & \mydots  & $\vert$ $t_{0,n}$ \textbf{X$_{h_{r,0}}$} & $\vert$ B(\textbf{N}, X$_{i_{0,1}}$) \textbf{X$_{j_{0,1}}$} & \mydots  & $\vert$B(\textbf{N}, X$_{i_{0,m}}$) \textbf{X$_{j_{0,m}}$}& $\vert$ R(\textbf{N}) \textbf{X$_{k_{0,1}}$}  & \mydots  & $\vert$R(\textbf{N}) \textbf{X$_{k_{0,p}}$} \\
        \textbf{\large X$_1$} & $\rightarrow$ &$\epsilon$ & $\vert$ $t_{1,1}$ \textbf{X$_{h_{1,1}}$}  & \mydots  & $\vert$ $t_{1,n}$ \textbf{X$_{h_{r,1}}$} & $\vert$ B(\textbf{N}, X$_{i_{1,1}}$) \textbf{X$_{j_{1,1}}$}  & \mydots  & $\vert$B(\textbf{N}, X$_{i_{1,m}}$) \textbf{X$_{j_{1,m}}$}& $\vert$ R(\textbf{N}) \textbf{X$_{k_{1,1}}$}  & \mydots  & $\vert$R(\textbf{N}) \textbf{X$_{k_{1,p}}$}\\
 &&&&&&& \multicolumn{1}{c}{  \mydots  } &&&&&\\
     \textbf{\large X$_r$} & $\rightarrow$ &$\epsilon$ & $\vert$ $t_{r,1}$ \textbf{X$_{h_{r,1}}$}  & \mydots  & $\vert$ $t_{r,n}$ \textbf{X$_{h_{r,n}}$} & $\vert$ B(\textbf{N}, X$_{i_{r,1}}$) \textbf{X$_{j_{r,1}}$}  & \mydots  & $\vert$B(\textbf{N}, X$_{i_{r,m}}$) \textbf{X$_{j_{r,m}}$}& $\vert$ R(\textbf{N}) \textbf{X$_{k_{r,1}}$}  & \mydots  & $\vert$R(\textbf{N}) \textbf{X$_{k_{r,p}}$}\\
     \textbf{\large N} & $\rightarrow$ &0 & $\vert$ 1  & \mydots  & $\vert$ n & $\vert$ n+1  & \mydots  & $\vert$ n+m& $\vert$ n+m+1  & \mydots  & $\vert$n+m+p
    \end{tabular}
    \caption{Grammar of \selfies, with recursion and \textbf{S}$\rightarrow$\textbf{X}$_0$.}              
    \label{tab:GrammarAbstract}
\end{table*}
The edge-multiplicity $\mu=\min(j,\gamma)$ is responsible for the semantic constraint of local degrees being satisfied. This is the most immediate constraint in many applications for physical sciences, which allows for 100\% validity. More complex, non-local constraints could be implemented by more complex grammars, such as explicit context-sensitive type-1 grammars.

\section{Potential applications to other domains in the physical sciences}

\begin{figure}[b]
    \centering
    \includegraphics[width=0.66\textwidth]{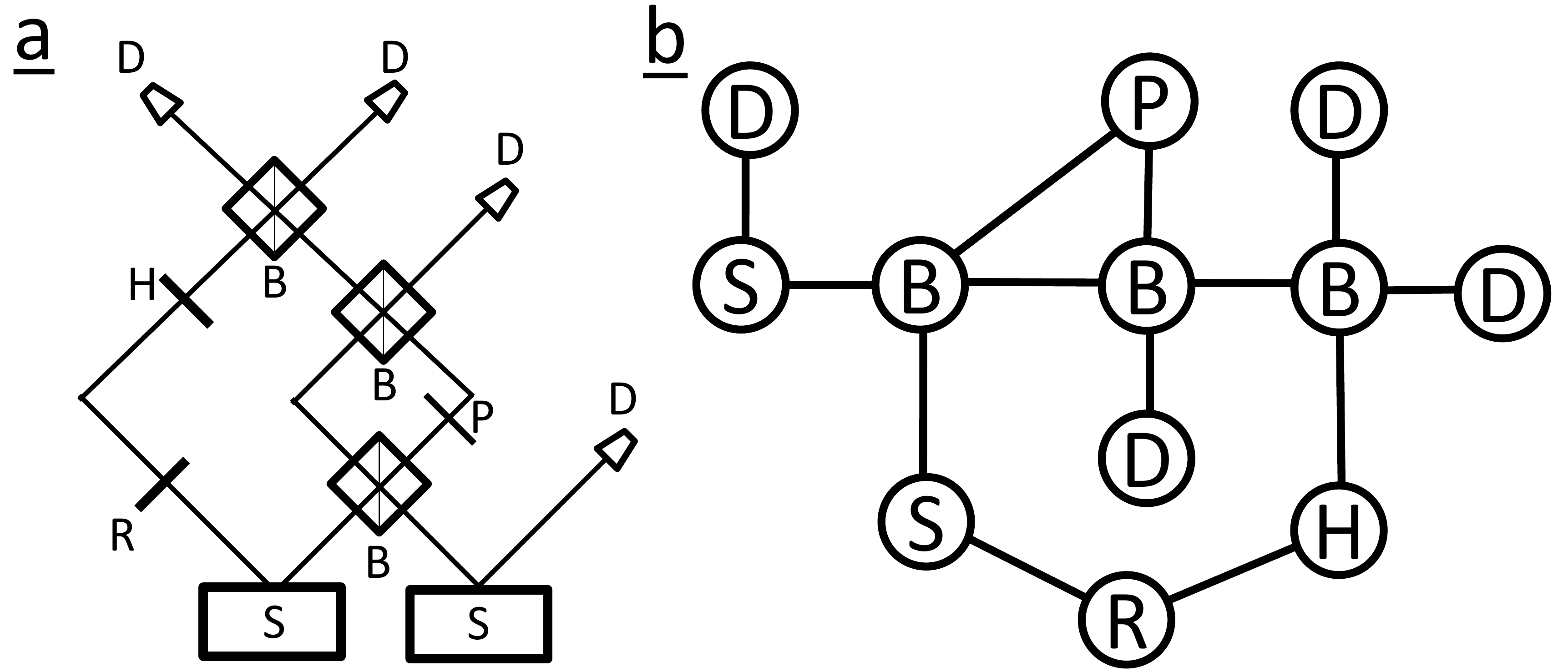}
    \caption{\selfies for quantum optical experiments. In (a) we see the graph generated from \selfies for a recent high-dimensional multipartite quantum experiment \cite{erhard2018experimental}. In (b), the structure of the experimental configuration.}
    \label{fig:quantumSI}
\end{figure}

\selfies can be used independently of the domain, which we demonstrate here. Ideal targets for \selfies grammar are different types of objects (which for the vertices) with vertex-dependent connectivity restrictions. In that case, rule vectors of grammars can be used to encode the restrictions on connectivities. Rings and Branches could be dependent on vertices as well. We now show now one different example from physics.

\subsection{Quantum Optical Experiments}
\setlength{\tabcolsep}{1pt}
\begin{table*}[t]
    \centering
    \begin{tabular}{llllllllll}
   &                                          & \textbf{\large S}                    & \multicolumn{1}{c}{\textbf{\large B}} & \multicolumn{1}{c}{\textbf{\large H}}          & \multicolumn{1}{c}{\textbf{\large P}}     & \multicolumn{1}{c}{\textbf{\large R}}    & \multicolumn{1}{c}{\textbf{\large D}}     & \multicolumn{1}{c}{\textbf{\large Y}}              & \multicolumn{1}{c}{\textbf{\large Z}}\\
        \textbf{\large X$_0$} & $\rightarrow$ & \texttt{$[$SPDC$]$} \textbf{X$_2$} & $\vert$ \texttt{$[$BS$]$} \textbf{X$_3$}    & $\vert$ \texttt{$[$Holo$]$} \textbf{X$_1$} & $\vert$ \texttt{$[$DP$]$} \textbf{X$_1$}  & $\vert$ \texttt{$[$Ref$]$} \textbf{X$_1$} & $\vert$ \texttt{$[$Det$]$}                & $\vert$ \textbf{X$_0$}                             & $\vert$ \textbf{X$_0$}   \\
        \textbf{\large X$_1$} & $\rightarrow$ & \texttt{$[$SPDC$]$} \textbf{X$_1$} & $\vert$ \texttt{$[$BS$]$} \textbf{X$_3$}    & $\vert$ \texttt{$[$Holo$]$} \textbf{X$_1$} & $\vert$ \texttt{$[$DP$]$} \textbf{X$_1$}  & $\vert$ \texttt{$[$Ref$]$} \textbf{X$_1$} & $\vert$ \texttt{$[$Det$]$}                & $\vert$ \textbf{X$_1$}                             & $\vert$ R(\textbf{N})\\
        \textbf{\large X$_2$} & $\rightarrow$ & \texttt{$[$SPDC$]$} \textbf{X$_1$} & $\vert$ \texttt{$[$BS$]$} \textbf{X$_3$}    & $\vert$ \texttt{$[$Holo$]$} \textbf{X$_1$} & $\vert$ \texttt{$[$DP$]$} \textbf{X$_1$}  & $\vert$ \texttt{$[$Ref$]$} \textbf{X$_1$} & $\vert$ \texttt{$[$Det$]$}                & $\vert$ B(\textbf{N},\textbf{X$_0$})\textbf{X$_1$} & $\vert$ R(\textbf{N})\textbf{X$_1$} \\ 
        \textbf{\large X$_3$} & $\rightarrow$ & \texttt{$[$SPDC$]$} \textbf{X$_1$} & $\vert$ \texttt{$[$BS$]$} \textbf{X$_3$}    & $\vert$ \texttt{$[$Holo$]$} \textbf{X$_1$} & $\vert$ \texttt{$[$DP$]$} \textbf{X$_1$}  & $\vert$ \texttt{$[$Ref$]$} \textbf{X$_1$} & $\vert$ \texttt{$[$Det$]$}                & $\vert$ B(\textbf{N},\textbf{X$_0$})\textbf{X$_2$} & $\vert$ R(\textbf{N})\textbf{X$_2$} \\ 
        \textbf{\large N}     & $\rightarrow$ &1                                   & $\vert$ 2                                   & $\vert$ 3                                  & $\vert$ 4                                 & $\vert$ 5                                & $\vert$ 6                                  & $\vert$ 8                                          & $\vert$ 9
    \end{tabular}
    \caption{Derivation rules of \selfies for a semantically restricted graph that represents quantum optical experiments, with the derivation starting in \textbf{X$_0$}.}              
    \label{tab:GrammarExampleQM}
\end{table*}  
A grammar for the generation of quantum optical experments can be written in Table \ref{tab:GrammarExampleQM}.

There, the non-terminal symbols stand for quantum optical components that are used in experiments, \texttt{$[$SPDC$]$} stands for a non-linear crystal that undergoes spontaneous parametric down-conversion to produce photon pairs, \texttt{$[$BS$]$} stands for beam splitters, \texttt{$[$Holo$]$} stands for holograms to modify the quantum state, \texttt{$[$DP$]$} stands for Dove prism which introduces mode dependent phases, \texttt{$[$Ref$]$} stand for mirrors which modify mode numbers and phases, and \texttt{$[$Det$]$} are single-photon detector. B(\textbf{N},\textbf{X$_0$}) and R(\textbf{N}) are branch functions and ring functions as defined in the main text. Now we derive a recent complex quantum optical experiment (which has been designed by a computer algorithm), which demonstrates high-dimensional multi-partite quantum entanglement \cite{erhard2018experimental}. The graph and the corresponding setup can be seen in Figure \ref{fig:quantumSI}.

\end{document}